\documentclass{article}

\usepackage{arxiv}

\usepackage[utf8]{inputenc}
\usepackage[T1]{fontenc}
\usepackage{amsmath,amssymb,amsfonts}
\usepackage{graphicx}
\usepackage{booktabs}
\usepackage{tabularx}
\usepackage{array}
\usepackage{subcaption}
\usepackage{xcolor}
\usepackage{wrapfig}
\usepackage{microtype}
\usepackage{url}
\newcolumntype{C}[1]{>{\centering\arraybackslash}p{#1}}
\usepackage[hidelinks]{hyperref}
\begin{document}
\title{CardiacMamba: Fair and Robust RGB-RF Fusion for Remote Heart Rate Estimation via State Space Modeling}
%
% 作者信息：Bo Zhao 与 Zheng Wu 为共同一作（*），Zitong Yu 为通讯作者（†），单位编号 1
\author{Bo Zhao$^{1*}$, Zheng Wu$^{1*}$, Yiping Xie$^{1}$, Zitong Yu$^{1\dagger}$ \\
  $^{1}$Great Bay University \\[0.4em]
  \footnotesize{$^{*}$These authors contributed equally to this work.\quad
  $^{\dagger}$Corresponding author.}
}
\maketitle
\begin{abstract}
Remote photoplethysmography (rPPG) enables non-contact heart rate (HR) monitoring from facial videos, but RGB-only methods are vulnerable to illumination changes, motion artifacts, and skin-tone-dependent optical reflectance. We propose CardiacMamba, a fair and robust RGB-RF fusion framework that integrates optical facial cues and radio-frequency cardiac motion cues through state space modeling. CardiacMamba introduces a Temporal Difference Mamba Module (TDMM) to enhance subtle RF temporal variations, a bidirectional SSM-based interaction mechanism to align heterogeneous RGB-RF dynamics, and a Channel-wise Fast Fourier Transform (CFFT) module for channel-domain spectral refinement. On the EquiPleth dataset, CardiacMamba achieves state-of-the-art performance with 0.96 bpm MAE, 3.06 bpm RMSE, and 0.97 Pearson correlation, while reducing the observed light-dark skin-tone MAE gap to 0.26 bpm and maintaining robustness under RGB degradation and RF-missing conditions.
\end{abstract}


\section{Introduction}
\label{sec:intr}

Remote photoplethysmography (rPPG) enables non-contact heart rate (HR) monitoring from facial videos, offering an unobtrusive alternative to contact-based ECG and PPG. Despite progress in physics-based~\cite{dehaan2013robust} and deep learning-based~\cite{yu2019remote} methods, RGB-based rPPG remains limited by its dependence on optical reflectance: illumination changes, motion artifacts, and reduced pulsatile contrast under darker skin tones degrade the signal-to-noise ratio.

Radio Frequency (RF) sensing offers a complementary mechanism. By capturing minute chest-wall vibrations through electromagnetic reflections, RF signals are largely insensitive to ambient lighting and skin pigmentation, but suffer from lower spatial resolution and vulnerability to body motion and multipath interference~\cite{alizadeh2019remote}. These complementary properties motivate RGB-RF fusion: RGB provides rich facial appearance cues, while RF supplies illumination- and skin-tone-invariant mechanical cardiac information.

Effective RGB-RF fusion, however, remains challenging. Existing strategies often rely on shallow feature concatenation or late-stage fusion, insufficiently modeling the heterogeneous dynamics between optical BVP signals and RF-induced chest motion. Three issues remain underexplored: aligning RGB and RF features with temporal characteristics in a shared representation space, enhancing physiological components via frequency-domain interaction and mitigating demographic disparities caused by the optical dependence of RGB-based rPPG.

To address these challenges, we propose CardiacMamba, an SSM-based RGB-RF fusion framework for fair and robust remote HR estimation. It introduces a Temporal Difference Mamba Module (TDMM) to enhance subtle RF temporal variations, a bidirectional SSM-based interaction mechanism to align heterogeneous RGB-RF dynamics, and a Channel-wise Fast Fourier Transform (CFFT) module for channel-domain spectral refinement. On EquiPleth, CardiacMamba achieves state-of-the-art accuracy, reduces skin-tone-related performance disparities, improves robustness under RGB degradation and RF-missing conditions.

% \begin{figure}
% \centering
% \begin{subfigure}{0.5\textwidth}
%     % \centering
%     \includegraphics[width=\linewidth, height=3cm]{Figures/3RGB.pdf}
%     \vspace{-2.0em}
%     \caption{RGB-only Methods}
%     \label{fig:rgb}
% \end{subfigure}
% \hfill
% \begin{subfigure}{0.5\textwidth}
%     \centering
%     \includegraphics[width=\linewidth, height=3cm]{Figures/RF3.pdf}
%     \vspace{-2.0em}
%     \caption{RF-only Methods}
%     \label{fig:rf}
% \end{subfigure}
% \hfill
% \begin{subfigure}{0.5\textwidth}
%     \centering
%     \includegraphics[width=\linewidth, height=3cm]{Figures/Fusion.pdf}
%     \vspace{-2.0em}
%     \caption{RGB-RF fusion Methods}
%     \label{fig:rgb_rf}
% \end{subfigure}
% \caption{Comparison of deep learning methods for rPPG learning. (a) RGB-only Method: Training with only RGB data collected by the camera. (b) RF-only Method: Training with only RF data collected by the radar. (c) Training with both RGB and RF data. }
% \label{fig:all_cameras}
% \vspace{-1.0em}
% \end{figure}
Our main contributions are summarized as follows:
\vspace{-0.5em}
\begin{itemize}
    \item We propose CardiacMamba, a state-space RGB-RF fusion framework that jointly exploits optical facial cues and illumination-invariant RF cardiac cues for robust and fair remote HR estimation.

    \item We develop a dynamic multimodal architecture where TDMM enhances RF temporal variations, a bidirectional SSM-based mechanism aligns heterogeneous RGB-RF dynamics, and CFFT performs channel-wise frequency-domain refinement.

    \item Extensive experiments on EquiPleth show that CardiacMamba achieves 0.96 bpm MAE, 3.06 bpm RMSE, and 0.97 Pearson correlation, reducing the observed light-dark skin-tone MAE gap to 0.26 bpm while maintaining robustness under RGB degradation and RF-missing conditions.
\end{itemize}
\vspace{-1em}
\section{Related Work}
\vspace{-1em}
\subsection{RGB Video-Based Methods}
\vspace{-0.5em}
RGB video-based rPPG estimates physiological signals from subtle facial appearance variations. Early methods relied on hand-crafted decomposition such as PCA and ICA~\cite{poh2010noncontact}, while deep learning later advanced rPPG with CNN~\cite{yu2019remote} and Transformer~\cite{yu2022physformer,yu2023physformerpp} architectures. Nevertheless, RGB-based rPPG remains constrained by optical sensing mechanism, motivating  non-visual modalities. Beyond physiological measurement, deep learning has also advanced face-video understanding in related tasks such as face anti-spoofing~\cite{yu2023dlfas,yu2021nasfas,yu2024rethinking,lin2025reliable,cai2025rehearsal,lin2026provable}, where robustness to appearance variation and multimodal cues is likewise essential.

\vspace{-1em}
\subsection{RF Radar-Based Methods}
\vspace{-0.5em}
RF radar measures minute chest displacements caused by cardiac motion; early frequency-domain pipelines~\cite{alizadeh2019remote} gave way to deep learning approaches~\cite{ZHANG2023105360}. RF is  robust to illumination and skin pigmentation, but has lower spatial resolution and is affected by body motion, so it is best used as a complement to RGB.

\vspace{-1em}
\subsection{Multi-modal Fusion Methods}
\vspace{-0.5em}
Prior work fused RGB with Infrared signals, and Vilesov et al.~\cite{vilesov2022blending} studied RGB-RF fusion with camera and 77 GHz radar. However, RGB videos and RF signals capture different manifestations of cardiac activity (optical blood-volume variations vs.~mechanical chest-wall motion), making simple concatenation or late fusion insufficient; existing methods often lack explicit cross-modal temporal alignment and frequency-domain interaction.
\vspace{-1em}
\subsection{Mamba and State Space Models}
\vspace{-0.5em}
State Space Models (SSMs)~\cite{gu2022efficiently} model long sequences through structured state transitions, and Vision Mamba (Vim)~\cite{zhu2024vision} introduces bidirectional state-space modeling with lower cost than Transformers, well suited to dense video or radar sequences with weak quasi-periodic dynamics. We build on these strengths for RGB-RF fusion. In parallel, recent works explore large language models~\cite{xie2026physllm}, physics-grounded harmonic attention~\cite{zhao2026phase}, multi-agent frameworks~\cite{yang2026physagent}, optimal-transport feature warping~\cite{zhao2026flow}, causal self-supervised learning~\cite{niu2026intervention}, and dual-branch structured attention~\cite{cao2026physnext} for remote physiological measurement; LLM-based multimodal understanding is also advancing rapidly across affective computing, e.g., audio-visual reasoning~\cite{ye2025cat}, micro-expression action units~\cite{liu2026aullm,liu2026aullmpp}, and emotion world models~\cite{wang2026affectagent,zhao2026affectverse}.

% \vspace{-0.3cm}
\vspace{-0.5em}
\section{Methodology}
\vspace{-0.5em}
\subsection{Preliminaries}
We adopt the discretized state space model (Mamba)~\cite{gu2022efficiently} for sequence modeling; the formal definitions of the continuous SSM, its discretization, and the convolutional equivalence are given in Appendix~\ref{app:ssm}.
\vspace{-1em}
\subsection{Overview}
\vspace{-0.5em}
\begin{figure*}[t]
\centering
\vspace{-1.5em}
\includegraphics[width=1\linewidth]{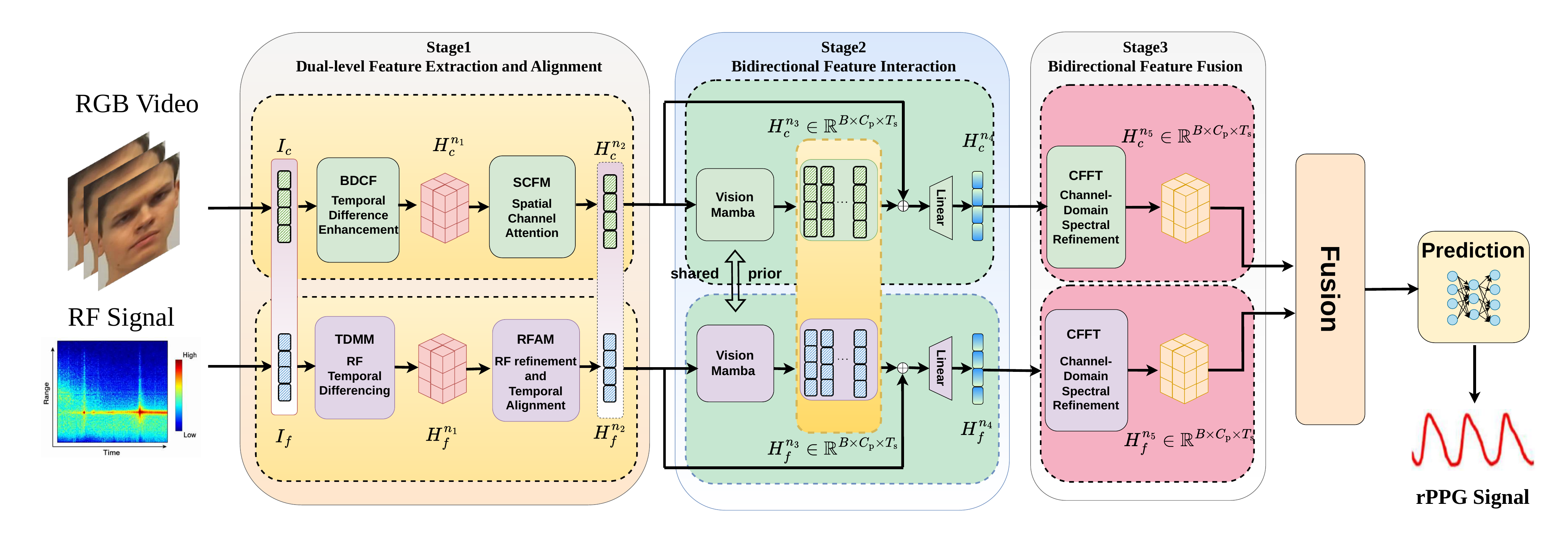}
\vspace{-2.0em}
  \caption{The overall architecture of CardiacMamba. It consists of three stages: Dual-level Feature Extraction and Alignment, Bidirectional Feature Interaction, and Bidirectional Feature Fusion.}
  \label{fig:network}
  \vspace{-2.0em}
\end{figure*}
As shown in Fig.~\ref{fig:network}, CardiacMamba estimates physiological signal by integrating complementary observations. Given an RGB video $I_{C}\in\mathbb{R}^{3\times T_{1}\times H\times W}$ and an RF input $I_{f}\in\mathbb{R}^{C\times T_{2}}$, the framework contains three stages: dual-level feature extraction, SSM-based temporal interaction and frequency-domain fusion.

First, modality-specific encoders extract physiological representations from the two streams. In the RGB branch, BDCF enhances BVP-related temporal color variations, while SCFM aggregates informative spatial-channel responses. In the RF branch, TDMM captures cardiac-induced temporal variations and two RFAMs refine RF features while aligning temporal resolution with RGB stream:
\begin{equation}
H_{c}^{n_{2}} = \mathrm{SCFM}(\mathrm{BDCF}(I_{C})),\qquad
H_{f}^{n_{2}} = \mathrm{RFAM}(\mathrm{RFAM}(\mathrm{TDMM}(I_{f}))).
\label{eq:dual_feature}
\end{equation}

\noindent Second, Vision Mamba (Vim) is used to model long-range temporal dependencies in both modalities under a 
\begin{figure}[t]
\centering
\includegraphics[width=\columnwidth]{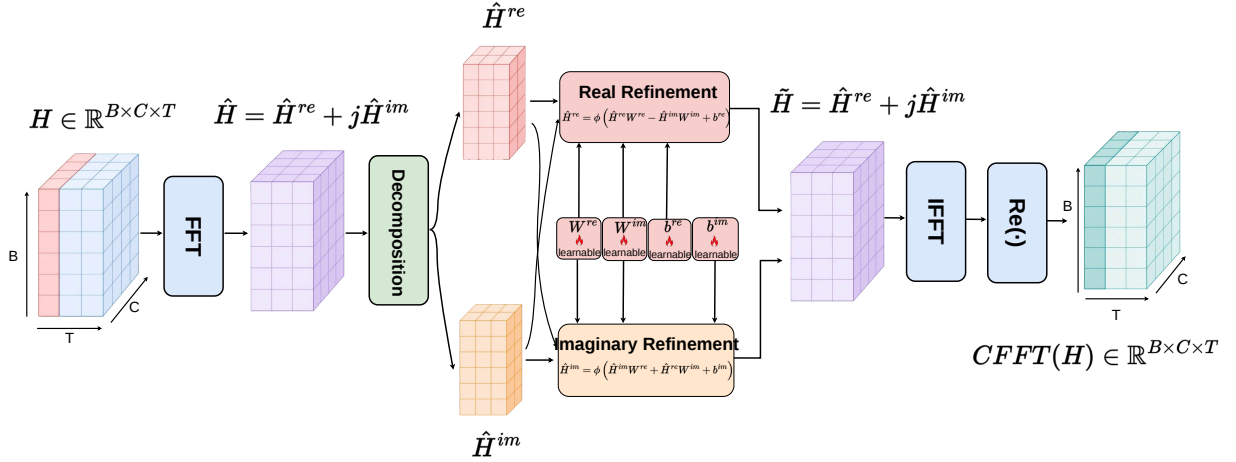}
\vspace{-2em}
\caption{Channel-wise Fast Fourier Transform (CFFT) for refining RGB and RF representations through channel-domain spectral interaction.}
\label{fig:cfft}
\vspace{-2em}
\end{figure}
% \vspace{-1em}
\noindent shared SSM-based dynamic prior. Residual connections and linear projections stabilize the representation refinement:
\begin{equation}
H_{c}^{n_{4}} = \mathrm{Linear}_{c}(H_{c}^{n_{2}}+\mathrm{Vim}_{c}(H_{c}^{n_{2}})),\qquad
H_{f}^{n_{4}} = \mathrm{Linear}_{f}(H_{f}^{n_{2}}+\mathrm{Vim}_{f}(H_{f}^{n_{2}})).
\label{eq:vim_encoding}
\end{equation}
This stage encourages RGB and RF features to share a coherent temporal evolution while preserving modality-specific cues. Finally, the Channel-wise Fast Fourier Transform (CFFT) module refines the channel spectra of both branches (Appendix~\ref{app:cfft}), and the refined representations are fused by the prediction head to reconstruct the BVP waveform:
\begin{equation}
\hat{y}= \mathrm{Predictor}(\mathrm{Fuse}(H_{c}^{n_{5}},H_{f}^{n_{5}})),
\label{eq:prediction}
\end{equation}
where $\mathrm{Fuse}(\cdot)$ denotes multimodal aggregation and $\hat{y}$ is the predicted signal.

\vspace{-0.5em}
\subsection{Dual-level Feature Extraction and Alignment}
\vspace{-0.5em}
\subsubsection{Low-level Feature Extraction}

\begin{figure}[t]
\centering
\includegraphics[width=\columnwidth,page=2]{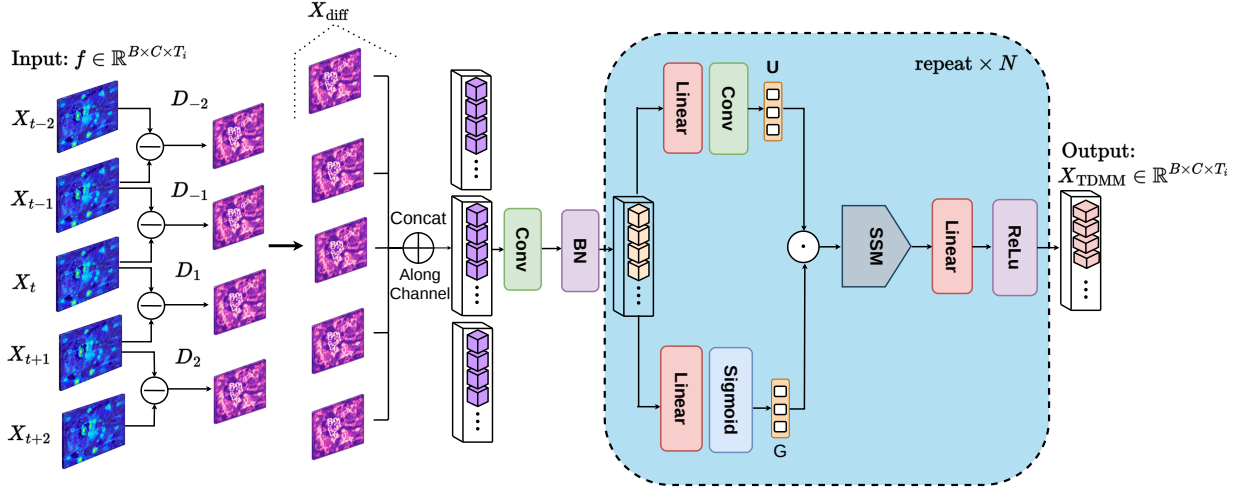}
\vspace{-0.4em}
\caption{Temporal Difference Mamba Module (TDMM) for extracting RF dynamic temporal features.}
\label{fig:tdmm}
\vspace{-2em}
\end{figure}

\textbf{Temporal Difference Mamba Module (TDMM).}
RF signals reflect cardiac activity through subtle chest-wall displacement, but such weak temporal variations are often buried by static reflections and low-frequency body motion. TDMM (Fig.~\ref{fig:tdmm}) highlights local temporal changes by frame differencing (Appendix~\ref{app:modules}) and captures long-range dependencies with an SSM-based Mamba block,
% \vspace{-0.5em}
whose output is computed as
\begin{equation}
\begin{aligned}
X_{\mathrm{TDMM}} &= \operatorname{ReLU}\big(\operatorname{Linear}_{o}(\operatorname{SSM}(G \odot \operatorname{Conv}_{7 \times 1}(\operatorname{Linear}_{u}(X_0))))\big),\\
G &= \sigma(\operatorname{Linear}_{g}(X_0)).
\end{aligned}
\label{eq:tdmm_output}
\end{equation}

\textbf{Bifurcated Diff-Conv Fusion (BDCF).}
For the RGB branch, BDCF enhances weak BVP-related color variations by processing the raw sequence and its temporal difference representation in parallel branches. The temporal differences follow the formulation in Appendix~\ref{app:modules}. The two branches are fused as
\begin{equation}
\begin{aligned}
X_{\mathrm{fu}} &= \alpha \operatorname{Stem}_{2}(X_{\mathrm{ori}}) + \beta \operatorname{Stem}_{2}(\alpha X_{\mathrm{ori}}+\beta X_{\mathrm{diff}}),\\
X_{\mathrm{ori}} &= \operatorname{Stem}_{1}(X),\quad X_{\mathrm{diff}} = \operatorname{Stem}_{1}(\operatorname{Concat}(D_{-2},D_{-1},D_{1},D_{2})).
\end{aligned}
\label{eq:bdcf}
\end{equation}
where $\operatorname{Stem}_{1}$ consists of a $7 \times 7$ convolution, batch normalization, ReLU activation, and max pooling; $\operatorname{Stem}_{2}$ consists of a $7 \times 7$ convolution, batch normalization, and ReLU activation. We set $\alpha=\beta=0.5$.

\vspace{-0.5em}
\subsubsection{High-level Feature Extraction}
\vspace{-0.5em}
\textbf{Spatial-Channel Fusion Module (SCFM).}
SCFM suppresses uninformative spatial responses and produces compact high-level RGB features: a lightweight $5 \times 5$ convolutional stem generates a normalized spatial attention (Appendix~\ref{app:modules}), and the attended representation is globally aggregated and refined as

The attended representation is then globally aggregated and refined:
\begin{equation}
X_{\mathrm{stem}} =
\operatorname{BN}
\left(
\operatorname{Conv}
\left(
\operatorname{GAP}(X_{\mathrm{attn}})
\right)
\right).
\label{eq:scfm_output}
\end{equation}

\textbf{RF Alignment Module (RFAM).}
RFAM refines RF features and aligns their temporal resolution with the RGB branch through local temporal convolution, channel attention, and strided downsampling (Appendix~\ref{app:modules}). Its output is
\begin{equation}
X_{\mathrm{RFAM}} =
\operatorname{ReLU}
\left(
\operatorname{Conv}_{7 \times 1}^{s=2}
\left(
\widetilde{X}_r
\right)
\right).
\label{eq:rfam_output}
\end{equation}
Through temporal refinement, channel reweighting, and resolution alignment, RFAM produces RF features compatible with subsequent RGB-RF fusion.

{
\color{black}
\vspace{-0.5em}
\subsection{Bidirectional Feature Fusion}
\vspace{-0.5em}
After modality-specific temporal modeling, the branches produce representations $H_{c}^{n_{4}}$ and $H_{f}^{n_{4}}$ that still contain modality-specific noise and redundant channels. We refine them with the Channel-wise Fast Fourier Transform (CFFT) module (Fig.~\ref{fig:cfft}), which performs spectral mixing along the feature-channel dimension: given $H \in \mathbb{R}^{B \times C \times T}$, a learnable complex-valued transformation refines the channel-frequency spectrum, and the real part of the inverse transform is retained as the output (Appendix~\ref{app:cfft}):
\begin{equation}
\operatorname{CFFT}(H)
=
\operatorname{Re}(H').
\label{eq:cfft_output}
\end{equation}
CFFT is applied to both modality branches:
\begin{equation}
H_{c}^{n_{5}} = \operatorname{CFFT}(H_{c}^{n_{4}}),
\qquad
H_{f}^{n_{5}} = \operatorname{CFFT}(H_{f}^{n_{4}}).
\label{eq:cfft_rgb_rf}
\end{equation}

For bidirectional multimodal fusion, each modality modulates the other through a lightweight cross-gating mechanism:
\begin{equation}
\widetilde{H}_{c} = H_{c}^{n_{5}} \odot \sigma(\operatorname{Linear}_{f \rightarrow c}(H_{f}^{n_{5}})),\qquad
\widetilde{H}_{f} = H_{f}^{n_{5}} \odot \sigma(\operatorname{Linear}_{c \rightarrow f}(H_{c}^{n_{5}})),
\label{eq:cross_modal_gate}
\end{equation}

\noindent where $\sigma(\cdot)$ is the Sigmoid function and $\odot$ denotes element-wise multiplication. The  refined features are aggregated and passed to the prediction head:
\begin{equation}
\widehat{y} = \operatorname{Predictor}(\operatorname{Fuse}(\widetilde{H}_{c},\widetilde{H}_{f})).
\label{eq:fusion_prediction}
\end{equation}

\vspace{-1em}
\section{Experiment}\label{sec4}
\vspace{-0.5em}
\label{sec:experiment}

The specifics of the datasets, evaluation metrics, and experimental setup, covering data splits, metric formulations are given in Appendices~\ref{app:datasets} and~\ref{app:setup}.
\vspace{-1em}
\subsection{Comparison with State-of-the-Art Methods}
\vspace{-0.5em}
\begin{table*}[!t]   % [!t] 强制置顶
\centering
\renewcommand{\arraystretch}{1.3}
\small
\setlength{\tabcolsep}{14pt}  % 可保留或去掉，用 \extracolsep 控制更均匀
\caption{Comparison of different methods on the EquiPleth dataset. The best results are marked in \textbf{bold}.}
\label{tab:comparison}
\begin{tabular*}{\linewidth}{@{\extracolsep{\fill}} l|cccc @{}}
\hline
Method & Input & MAE & RMSE & $\rho$ \\ \hline
DeepPhys~\cite{chen2018deepphys} & RGB & 5.54 & 18.51 & 0.66 \\ 
PhysNet~\cite{yu2019remote} & RGB & 8.06 & 19.71 & 0.61 \\ 
MTTS-CAN ~\cite{liu2020multi} & RGB & 3.69 & 13.8 & 0.82 \\ 
PhysFormer~\cite{yu2022physformer} & RGB & 12.92 & 24.36 & 0.47 \\ 
EfficientPhys~\cite{liu2023efficientphys} & RGB & 5.47 & 17.04 & 0.71 \\ 
RhythmMamba~\cite{zou2024rhythmmamba} & RGB & 2.87 & 9.58 & 0.92 \\ 
\textbf{Ours (RGB-Only)} & RGB & 1.2 & 4.23 & 0.95 \\ \hline
Tu et al. & RF & 5.5 & 11.68 & 0.64 \\ 
Mercuri et al. & RF & 4.73 & 9.6 & 0.7 \\  
FFT-based \cite{alizadeh2019remote} & RF & 13.51 & 21.07 & 0.24 \\ 
\textbf{Ours (RF-Only)} & RF & 5.2 & 7.4 & 0.8 \\ \hline
Vilesov et al. \cite{vilesov2022blending} & RGB+RF & 1.12 & 3.42 & 0.95  \\ 
\textbf{Ours (Full model)} & RGB+RF & \textbf{0.96} & \textbf{3.06} & \textbf{0.97} \\ 
\hline
\end{tabular*}
\vspace{-1em}
\end{table*}
Table~\ref{tab:comparison} compares CardiacMamba with representative RGB-only, RF-based, and RGB-RF multimodal baselines~\cite{vilesov2022blending}. CardiacMamba achieves the best overall performance, with 0.96 bpm MAE, 3.06 bpm RMSE, and 0.97 Pearson correlation. Compared with the best RGB-only baseline, it reduces MAE and RMSE by 66.6\% and 68.1\%, respectively; compared with Vilesov et al.~\cite{vilesov2022blending}, it further reduces MAE and RMSE by 14.3\% and 10.5\%. These results validate the effectiveness of integrating RF cues with RGB features through TDMM, SSM-based temporal modeling, and CFFT-based channel-domain refinement. In the RF-only setting, our RF branch achieves the lowest RMSE and highest correlation among RF-based methods, although its MAE remains slightly higher than the strongest RF-only baseline, indicating that RF is most effective as a complementary modality rather than a replacement for RGB.
\vspace{-1.5em}
\subsection{Measuring Skin Tone Bias and Fairness}
\vspace{-0.5em}
We assess skin-tone fairness by measuring light-dark group performance gaps in Table~\ref{tab:comparisonb}, where smaller gaps indicate more consistent HR estimation across demographic groups. CardiacMamba achieves the smallest MAE gap of 0.26 bpm, reducing the gap by 61.2\% compared with Vilesov et al.~\cite{vilesov2022blending} and substantially outperforming ICA~\cite{poh2010advancements} and PhysNet~\cite{yu2019remote}, whose MAE gaps are 4.42 bpm and 2.22 bpm, respectively. It also obtains a 1.28 bpm RMSE gap and a 0.05 Pearson correlation gap, lower in magnitude than those of most RGB-based baselines~\cite{yu2019remote}. These results suggest that RGB-RF fusion mitigates skin-tone-related imbalance by reducing reliance on optical skin reflectance, while fairness is reported as an observed EquiPleth performance disparity.

\vspace{-1em}
\subsection{Measurement in Missing Modality Scenarios}
\vspace{-0.5em}
\begin{table*}[!t]
\centering
\renewcommand{\arraystretch}{1.1}
\small
\setlength{\tabcolsep}{5pt}
\caption{Testing under missing-modality conditions.}
\label{tab:comparison3}
\begin{tabular*}{\linewidth}{@{\extracolsep{\fill}} llllll @{}}
\hline
Method & Train & Test & MAE & RMSE  \\ 
\midrule
Base & RGB\&RF & RGB & 21.7 & 25.7  \\ 
Base & RGB\&RF & RF & 21.4 & 24.3  \\ 
Base & RGB\&RF & RGB\&RF & 20.3 & 24.8  \\ \hline
Vilesov et al.~\cite{vilesov2022blending} & RGB\&RF & RGB & 6.82 & 13.32  \\ 
Vilesov et al.~\cite{vilesov2022blending} & RGB\&RF & RF & \textbf{7.25} & \textbf{9.62}  \\ 
Vilesov et al.~\cite{vilesov2022blending} & RGB\&RF & RGB\&RF & 1.12 & 3.42  \\ \hline 
\textbf{CardiacMamba (Ours)} & RGB\&RF & RGB & \textbf{1.2} & \textbf{3.41}  \\ 
\textbf{CardiacMamba (Ours)} & RGB\&RF & RF & 11.0 & 13.0  \\ 
\textbf{CardiacMamba (Ours)} & RGB\&RF & RGB\&RF & \textbf{0.96} & \textbf{3.06}  \\ 
\bottomrule
\end{tabular*}
\vspace{-2em}
\end{table*}

Table~\ref{tab:comparison3} evaluates missing-modality robustness by training with RGB-RF inputs and testing with full or partial modalities. With both modalities available, CardiacMamba outperforms Vilesov et al.~\cite{vilesov2022blending}, reducing MAE and RMSE by 14.3\% and 10.5\%, respectively. When RF is absent, it maintains strong RGB-only inference with 1.20 bpm MAE, close to the full-modality result of 0.96 bpm, and reduces MAE by 82.4\% over Vilesov et al.~\cite{vilesov2022blending} under the same setting. However, when RGB is unavailable, performance degrades to 11.00 bpm MAE and 13.00 bpm RMSE, revealing asymmetric robustness. Thus, CardiacMamba is robust to RF missingness and RGB degradation, but RF-only

% fallback remains a limitation for future work.

% \newcolumntype{C}[1]{>{\centering\arraybackslash}p{#1}}

\vspace{-1em}
\subsection{Ablation Study}
\vspace{-0.5em}
\begin{table*}[!t]   % 跨栏并强制置顶；若单栏请改为 \begin{table}[!t]
    \vspace{-0.5em}
    \centering
    \small
    \caption{Ablation on different modules, including `Vim' (short for Vision Mamba), `CFFT' (short for Channel-wise Fast Fourier Transform), `RFAM' (short for RF Alignment Module), and `TDMM' (short for Temporal Difference Mamba Module). }
    \label{tab:comparison4}
    \renewcommand\arraystretch{1.0}
    \setlength{\tabcolsep}{1.3mm}
    \begin{tabular*}{\linewidth}{@{\extracolsep{\fill}} llllllll @{}}
        \toprule
        Vim & CFFT & SSM & RFAM & TDMM & MAE & RMSE & $\rho$ \\ 
        \midrule
        $\times$ & \checkmark & \checkmark & \checkmark & \checkmark & 1.7 & 4.9 & 0.91 \\ 
        \checkmark & $\times$ & \checkmark & \checkmark & \checkmark & 4.92 & 6.33 & 0.77 \\ 
        \checkmark & \checkmark & $\times$ & \checkmark & \checkmark & 1.86 & 5.51 & 0.89 \\ 
        \checkmark & \checkmark & \checkmark & $\times$ & \checkmark & 1.85 & 5.31 & 0.9 \\ 
        \checkmark & \checkmark & \checkmark & \checkmark & $\times$ & 3.82 & 8.06 & 0.81 \\
        \checkmark & \checkmark & \checkmark & \checkmark & \checkmark & \textbf{0.96} & \textbf{3.06} & \textbf{0.97}   \\
        \bottomrule
    \end{tabular*}
    \vspace{-1.5em}
\end{table*}
\begin{table*}[!t]   % 跨栏置顶
\centering
\small
\renewcommand{\arraystretch}{1.05}
\setlength{\tabcolsep}{4pt}
\caption{Ablation study on model robustness under simulated RGB signal degradation. We added Gaussian noise to the RGB input to test performance. Best results under noisy conditions are in \textbf{bold}.}
\label{tab:noise_ablation}
\begin{tabular*}{\linewidth}{@{\extracolsep{\fill}} l l c c c @{}}
    \toprule
    \textbf{Model} & \textbf{Condition}
    & \textbf{MAE}
    & \textbf{RMSE}
    & \textbf{$\rho$} \\
    \midrule
    \textbf{Ours (RGB-Only)} & Normal           & 1.20 & 4.23  & 0.95 \\
                             & + Gaussian Noise & 8.54 & 15.12 & 0.41 \\
    \midrule
    \textbf{Ours (Full model)} & Normal           & 0.96 & 3.06 & 0.97 \\
                               & + Gaussian Noise & \textbf{2.15} & \textbf{5.80} & \textbf{0.88} \\
    \bottomrule
\end{tabular*}
\vspace{-2em}
\end{table*}

\textbf{Impact of Key Modules.}
Table~\ref{tab:comparison4} evaluates the contributions of Vim, CFFT, SSM, RFAM, and TDMM. Removing 
any component degrades performance, confirming that each module benefits RGB-RF representation learning. CFFT is the most critical: without it, MAE increases from 0.96 bpm to 4.92 bpm and $\rho$ drops from 0.97 to 0.77, indicating that channel-domain spectral refinement is essential for stabilizing multimodal fusion. TDMM is also important for RF temporal modeling; removing it increases MAE and RMSE to 3.82 bpm and 8.06 bpm, respectively. Removing Vim raises MAE to 1.70 bpm, validating the benefit of SSM-based long-range temporal encoding, while removing SSM or RFAM leads to smaller but consistent degradation, with MAE increasing to 1.86 bpm and 1.85 bpm. These results show that CFFT and TDMM are  influential, while SSM and RFAM further improve temporal stability and RF alignment.

\textbf{Robustness to RGB Degradation.}
Table~\ref{tab:noise_ablation} evaluates robustness under Gaussian noise added to the RGB input. The RGB-only model degrades severely, with MAE increasing from 1.20 bpm to 8.54 bpm and $\rho$ dropping to 0.41. In contrast, the full RGB-RF model maintains substantially better performance under the same corruption, achieving 2.15 bpm MAE and 0.88 correlation. This demonstrates that RF provides a stable complementary signal when visual observations are degraded, supporting RGB-RF fusion for robust physiological measurement in challenging visual conditions.

\vspace{-1em}
\subsection{Visualization and Analysis}
\vspace{-0.5em}

\begin{figure*}[t]
\centering
\begin{subfigure}[t]{0.20\textwidth}
    \centering
    \includegraphics[width=\linewidth, height=3cm]{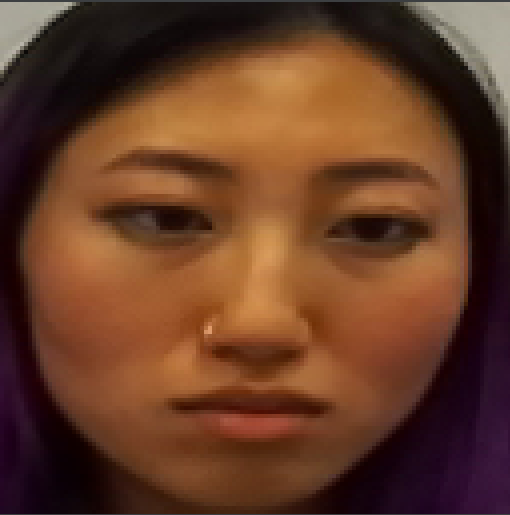}
    % \caption{Human face}
    \label{fig:face_raw1}
\end{subfigure}
\hfill
\begin{subfigure}[t]{0.20\textwidth}
    \centering
    \includegraphics[width=\linewidth, height=3cm]{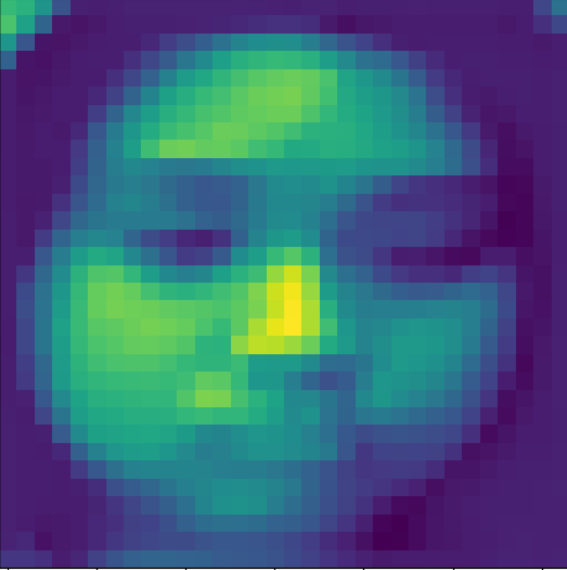}
    % \caption{Feature heat map of human face}
    \label{fig:face_af2}
\end{subfigure}
\hfill
\begin{subfigure}[t]{0.20\textwidth}
    \centering
    \includegraphics[width=\linewidth, height=3cm]{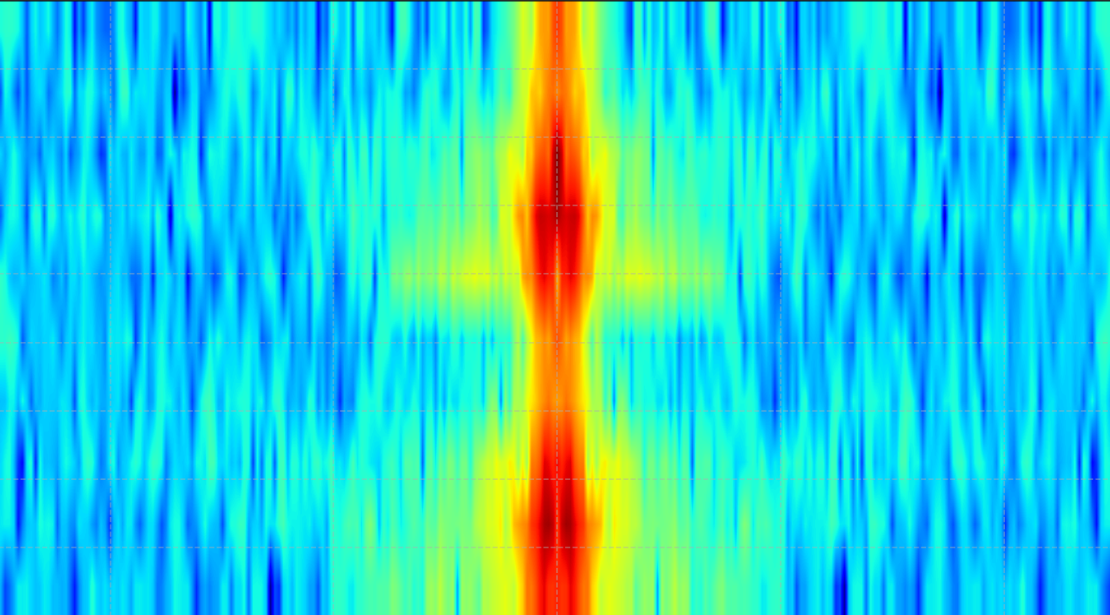}
    % \caption{Radar spectrum diagram}
    \label{fig:rf_raw3}
\end{subfigure}
\hfill
\begin{subfigure}[t]{0.20\textwidth}
    \centering
    \includegraphics[width=\linewidth, height=3cm]{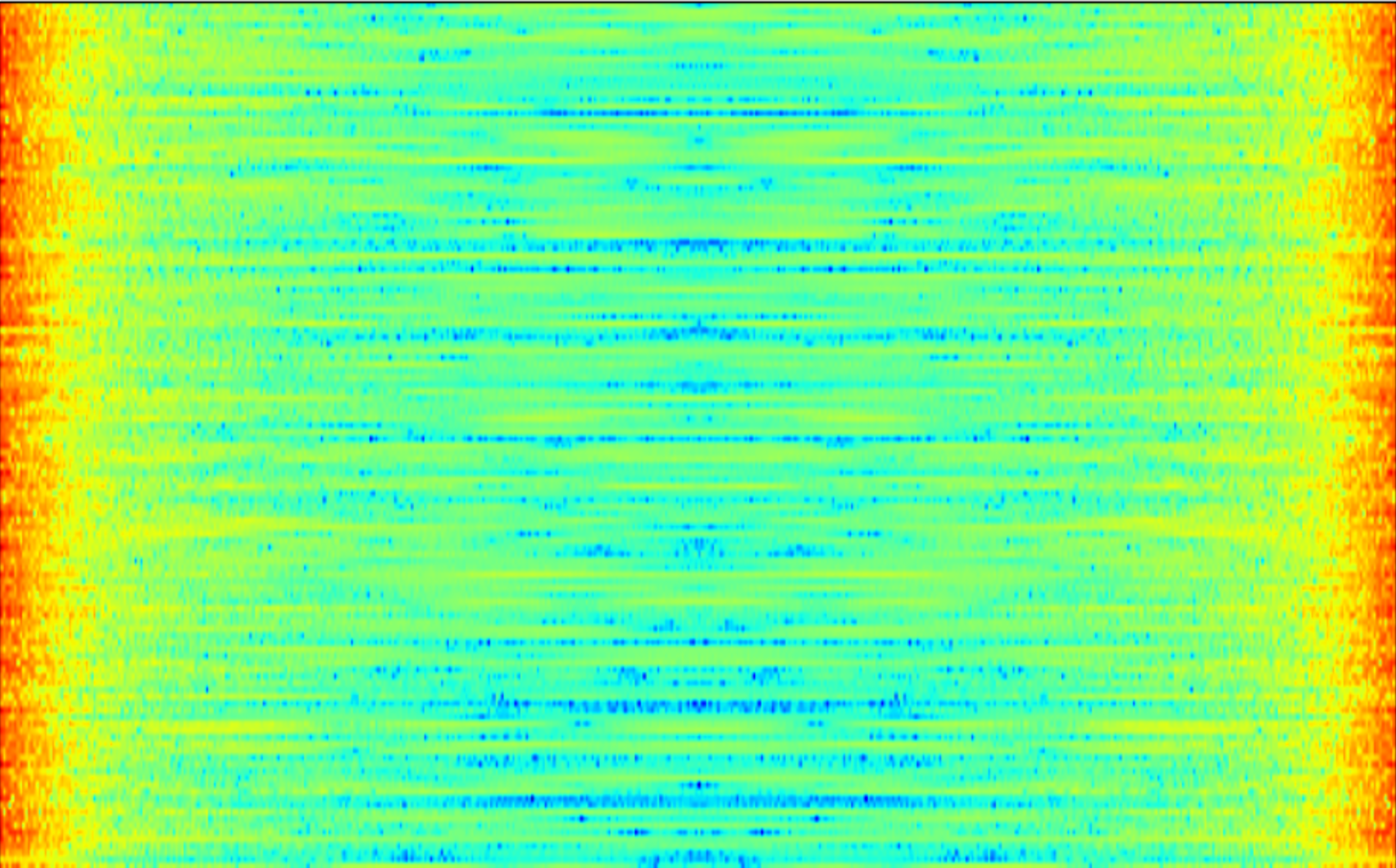}
    % \caption{Feature heat map of radar spectrum diagram}
    \label{fig:rf_af4}
\end{subfigure}
\vspace{-1em}
\caption{Visual representations of features (from left to right): human face, feature heat map, radar spectrum diagram, and its feature heat map.}
\label{fig:full_page_feature1}
\vspace{-2em}
\end{figure*}
Fig.~\ref{fig:full_page_feature1} visualizes the learned RGB and RF representations. Both modalities contribute distinct yet complementary 
\begin{wrapfigure}{r}{0.5\columnwidth}
% \vspace{-2em}
\centering
\captionsetup{justification=centering}

\begin{subfigure}[t]{0.45\linewidth}
    \centering
    % \vspace{-1em}
    \includegraphics[width=\linewidth,height=2.2cm,keepaspectratio]{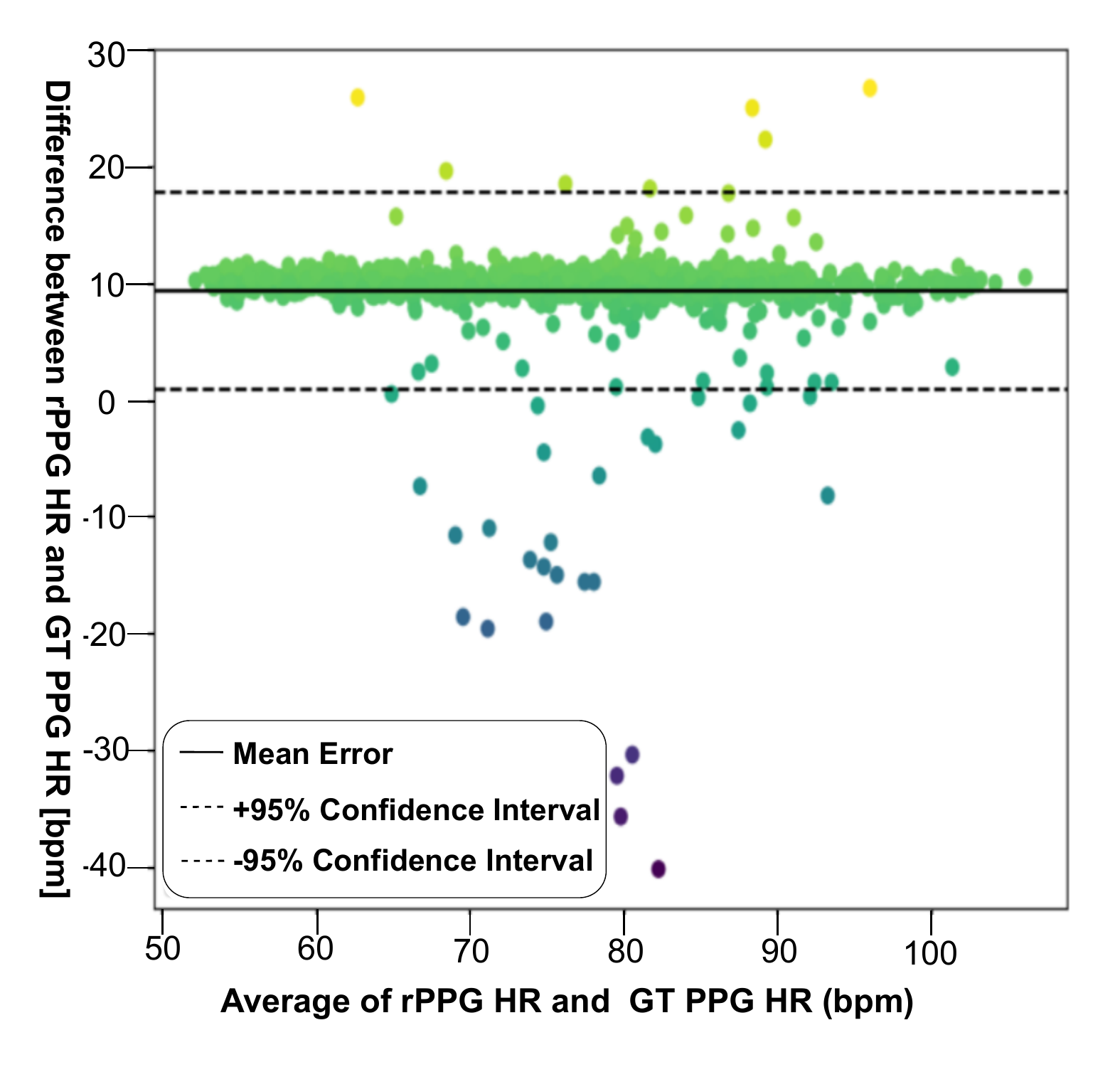}
    \caption{CardiacMamba}
    \label{fig:face_raw2}
\end{subfigure}
\hfill
\begin{subfigure}[t]{0.5\linewidth}
    \centering
    \includegraphics[width=\linewidth,height=2.2cm,keepaspectratio]{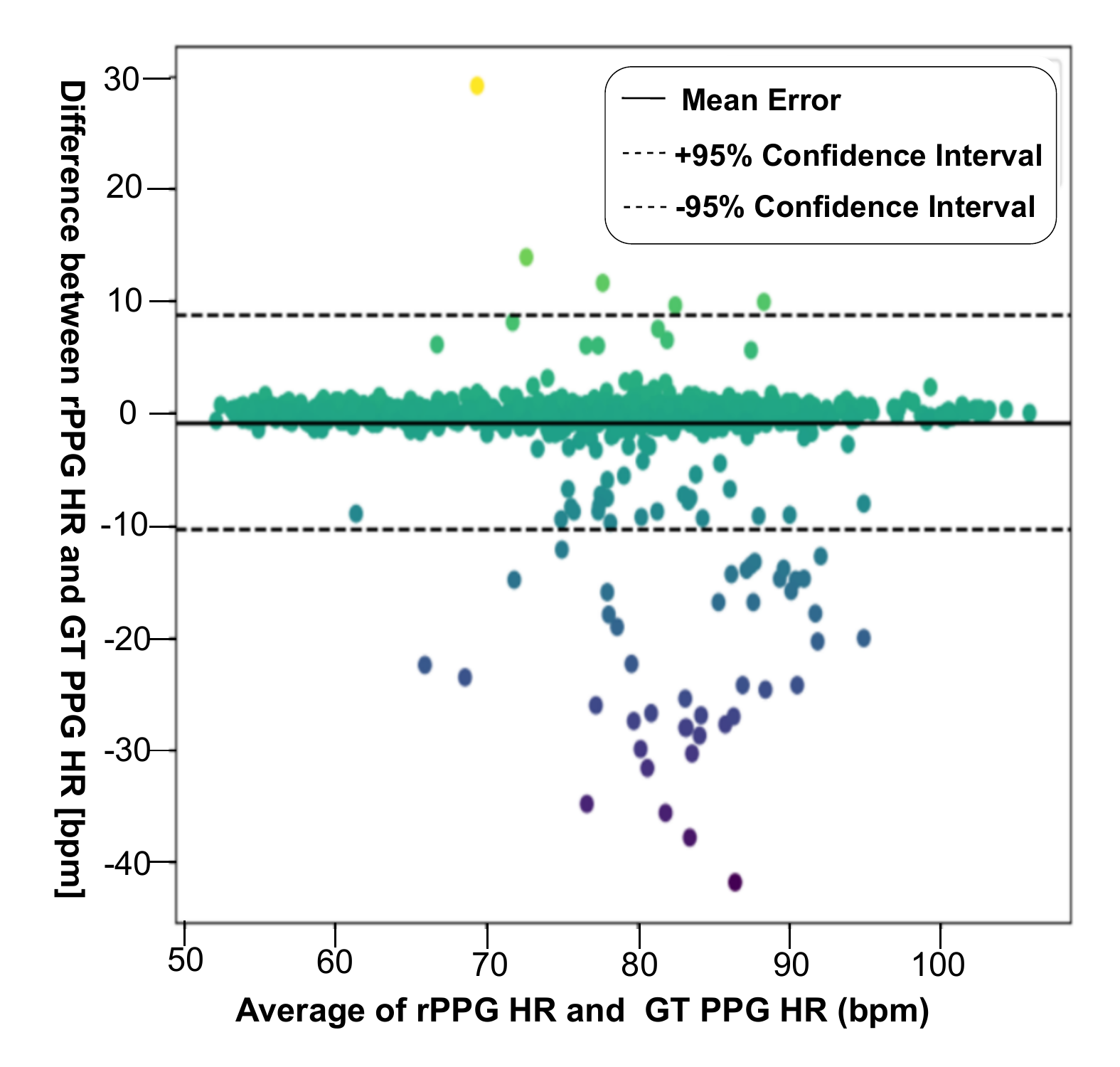}
    \caption{Vilesov et al.}
    \label{fig:face_af3}
\end{subfigure}

\captionsetup{justification=raggedright}
\caption{Bland-Altman plots comparing estimated and ground-truth heart rates.}
\label{fig:full_page_feature2}
% \vspace{0.4em}

\includegraphics[width=\linewidth,height=2.2cm,keepaspectratio]{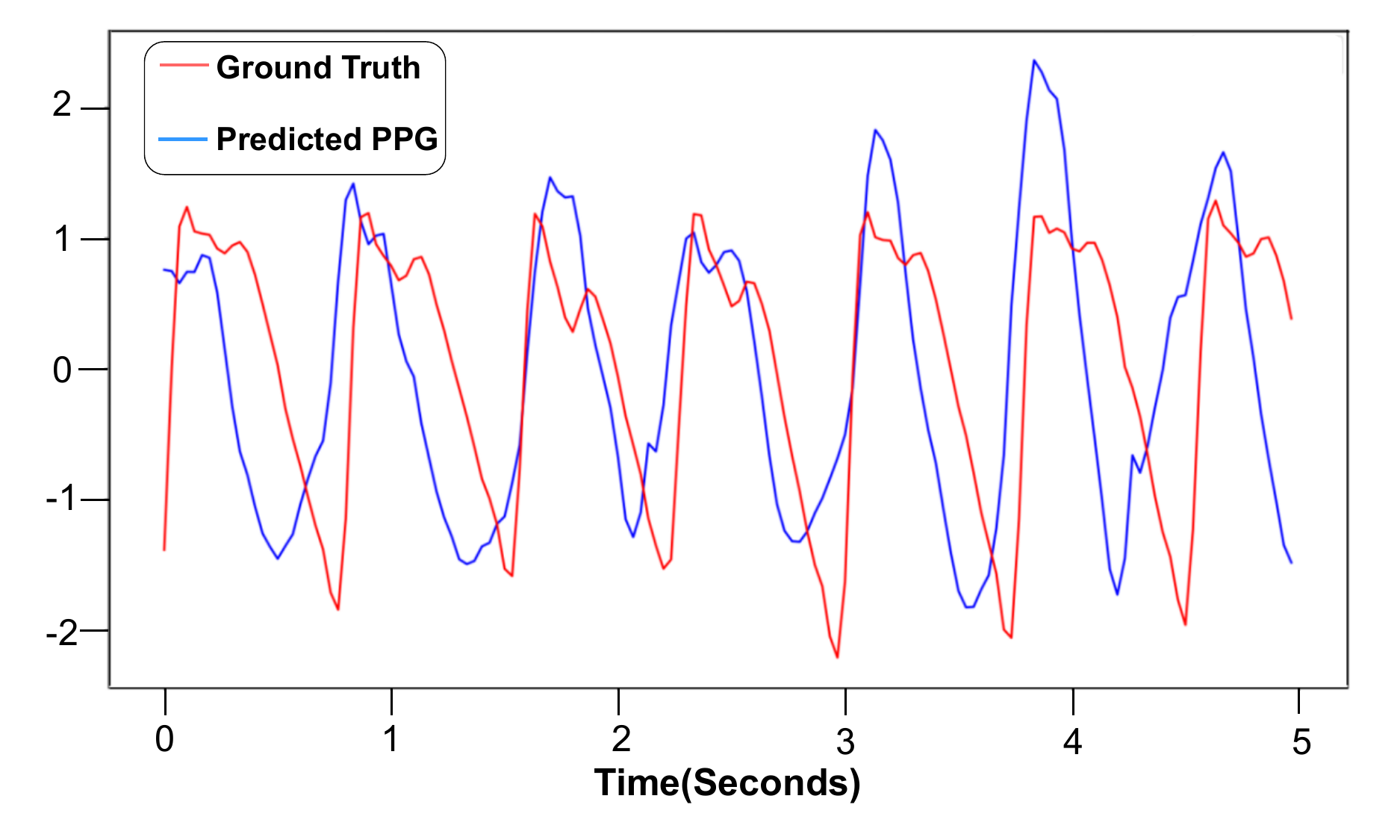}
\vspace{-0.5em}
\caption{Comparison of ground-truth and predicted PPG signals.}
\label{fig:face_af4}
\vspace{-3em}
\end{wrapfigure}
information: the optical branch encodes superficial blood-volume changes, whereas the RF branch reflects deeper mechanical cardiac activity, jointly supporting robust estimation under diverse conditions. The RGB heatmap 
highlights BVP-related facial regions, while the RF heatmap emphasizes radar responses associated with physiological motion, indicating that CardiacMamba captures optical and mechanical cardiac cues.

\textbf{Bland-Altman Analysis.}
Fig.~\ref{fig:full_page_feature2} compares HR agreement with ground 
truth. CardiacMamba exhibits tighter clustering within the confidence intervals than Vilesov et al.~\cite{vilesov2022blending}, showing improved estimation consistency.

\textbf{Signal Reconstruction Quality.}
Fig.~\ref{fig:face_af4} shows that the predicted PPG 
signal follows the ground-truth periodic pattern, providing qualitative evidence of physiological waveform recovery. This phase-consistent alignment further suggests that CardiacMamba preserves beat-to-beat temporal dynamics rather than merely matching the dominant heart-rate frequency.

% \begin{figure}[pos=h] % 尝试用 h 放在当前位置
%   \centering
%   % 关键：使用 \linewidth,它会自动适应单栏的宽度
%   \includegraphics[width=\linewidth]{imgs/yellow4.pdf} 
  
%   \vspace{-1.5em} % 如果觉得图跟下面文字太远,可以微调
%   \caption{Visualization of continuous HR results...}
%   \label{fig:4}
%   \vspace{-1.0em}
% \end{figure}
\vspace{-1em}
\section{Conclusion}
\vspace{-1em}
We introduced CardiacMamba, an RGB-RF fusion framework integrating TDMM, bidirectional SSM, and CFFT for robust and fair remote HR estimation. It achieves state-of-the-art accuracy on EquiPleth, reduces observed skin-tone-related disparities, and remains robust under RGB degradation and RF-missing conditions. Future work will address the limited RF-only fallback and validate generalization across larger populations.

\appendix
\section{Datasets and Metrics}\label{app:datasets}

\textbf{Datasets.}\quad
We evaluate CardiacMamba on EquiPleth~\cite{vilesov2022blending}, a synchronized RGB-RF benchmark for equitable remote physiological measurement. It contains recordings from 91 subjects, including 28 light-skin,

\begin{table}
    \color{black}
    \centering

    % \vspace{-0.3em} % 调整表格与标题的间距
    \renewcommand\arraystretch{1} % 调整行高
    \setlength{\tabcolsep}{2.5mm} % 调整列间距
    \caption{\color{black}Comparison of the Fairness of various methods on the RGB and RF fusion task, with the differences in performance ($\Delta$) between light and dark skin tones in terms of MAE, RMSE, and correlation coefficient ($\rho$). }
    \label{tab:comparisonb} % 添加表格标签
    \resizebox{0.48\textwidth}{!} { % 调整表格宽度
        \begin{tabular}{@{}lllll@{}}
            \toprule
            Method & Input & $\Delta$MAE & $\Delta$RMSE & $\Delta$$\rho$ \\ 
            \midrule
            ICA \cite{poh2010advancements} & RGB & 4.42 & 3.15 & -0.36 \\ 
            CHROM \cite{dehaan2013robust} & RGB & 4.97 & 4.17 & -0.38 \\ 
            BCG \cite{balakrishnan2013detecting}& RGB & 0.99 & \textbf{1.25} & \textbf{0.05} \\ 
            PhysNet \cite{yu2019remote} & RGB & 2.22 & 4.05 & -0.25 \\ 
            FFT-based RF \cite{alizadeh2019remote} & RF & 1.32 & 2.06 & 0.32 \\
           Vilesov et al. \cite{vilesov2022blending} & RGB\&RF & 0.67 & 1.44 & -0.10 \\ 
            \textbf{CardiacMamba (Ours)} & RGB\&RF & \textbf{0.26} & 1.28 & \textbf{0.05} \\ 
            \bottomrule
        \end{tabular}
    }
    % \vspace{-1.0em} % 调整表格与下文的间距
\end{table}

49 medium-skin, and 14 dark-skin subjects. Each subject participated in six 30-second sessions captured by an RGB camera at 30 fps and a 77 GHz FMCW radar. The paired streams provide complementary cardiac observations: RGB videos encode facial blood-volume-related appearance variations, while RF signals capture mechanical chest-wall motion.

We follow a subject-independent protocol using the predefined training, validation, and testing folds, ensuring that

no subject appears in more than one subset. This prevents identity leakage and enables reliable evaluation of cross-subject generalization.

\textbf{Metrics.}\quad
Heart rate estimation is evaluated using Mean Absolute Error (MAE), Root Mean Square Error (RMSE),
and Pearson correlation coefficient ($\rho$). MAE and RMSE are reported in beats per minute (bpm), where lower values indicate better accuracy, while higher $\rho$ indicates stronger agreement with ground-truth HR.

\section{Experimental Setup}\label{app:setup}
\subsection{Experimental Setup}
\label{sec:imp}

For RGB preprocessing, facial regions are detected by MTCNN~\cite{zhang2016joint}, cropped from each frame, resized to $128 \times 128$, converted to floating-point tensors, and normalized by 255. For RF preprocessing, raw IQ samples are transformed into range profiles using Discrete Fourier Transform (DFT) and stacked over time to form range-time representations. We select the range bin with the highest average energy and retain a 25 cm neighboring window to focus on torso-related cardiac displacement while suppressing background reflections. The RF inputs are normalized by $1.255 \times 10^{5}$, processed with IQ rotation, and reshaped into channel-temporal representations.

The model is trained for 30 epochs on an NVIDIA RTX 4090 GPU using Adam with a batch size of 32, an initial learning rate of $3 \times 10^{-4}$, and a weight decay of $1 \times 10^{-2}$. The best checkpoint is selected on the validation set and evaluated on the held-out test set. Unless otherwise specified, all learning-based baselines follow the same subject-independent protocol.

\section{State Space Model Preliminaries}\label{app:ssm}
State Space Models (SSMs) provide an efficient formulation for long-range sequence modeling through latent state evolution. Given a continuous input $x(t)\in\mathbb{R}$, an SSM maps it to an output $y(t)\in\mathbb{R}$ via a hidden state $h(t)\in\mathbb{R}^{N}$:
\begin{equation}
\begin{aligned}
\frac{d h(t)}{dt} &= \mathbf{A}h(t)+\mathbf{B}x(t), \\
y(t) &= \mathbf{C}h(t)+\mathbf{D}x(t),
\end{aligned}
\label{eq:ssm_continuous}
\end{equation}
where $\mathbf{A}$, $\mathbf{B}$, $\mathbf{C}$, and $\mathbf{D}$ denote the state transition, input projection, output projection, and skip connection parameters, respectively.

For neural sequence modeling, the continuous system is discretized with step size $\Delta$ using Zero-Order Hold:
\begin{equation}
\overline{\mathbf{A}}=\exp(\Delta\mathbf{A}),\qquad
\overline{\mathbf{B}}=(\Delta\mathbf{A})^{-1}
\left(\exp(\Delta\mathbf{A})-\mathbf{I}\right)\Delta\mathbf{B}.
\label{eq:ssm_discretization}
\end{equation}
The resulting recurrence is
\begin{equation}
h_k=\overline{\mathbf{A}}h_{k-1}+\overline{\mathbf{B}}x_k,\qquad
y_k=\mathbf{C}h_k+\mathbf{D}x_k ,
\label{eq:ssm_recurrence}
\end{equation}
which can be equivalently computed as a structured convolution:
\begin{equation}
\mathbf{y}=\overline{\mathbf{K}}*\mathbf{x}+\mathbf{D}\mathbf{x},\quad
\overline{\mathbf{K}}=
\left(
\mathbf{C}\overline{\mathbf{B}},
\mathbf{C}\overline{\mathbf{A}}\overline{\mathbf{B}},
\ldots,
\mathbf{C}\overline{\mathbf{A}}^{L-1}\overline{\mathbf{B}}
\right).
\label{eq:ssm_convolution}
\end{equation}
This state-evolution structure enables efficient modeling of weak and quasi-periodic physiological dynamics over long RGB and RF sequences.

\section{Channel-wise Fast Fourier Transform Details}\label{app:cfft}
Unlike temporal Fourier analysis that directly estimates physiological frequencies, CFFT performs spectral mixing along the feature-channel dimension. This design models inter-channel dependencies in a transformed feature space, allowing informative responses to be enhanced while noisy or redundant channels are suppressed. Given $H \in \mathbb{R}^{B \times C \times T}$, CFFT applies a discrete Fourier transform along the channel dimension:
\begin{equation}
\widehat{H}_{b,k,t}
=
\sum_{c=0}^{C-1}
H_{b,c,t}
\exp\left(
-j\frac{2\pi k c}{C}
\right),
\quad
k=0,\ldots,C-1,
\label{eq:cfft_dft}
\end{equation}
where $c$ and $k$ denote the channel index and channel-frequency index, respectively. The transformed feature is decomposed as
\begin{equation}
\widehat{H}
=
\widehat{H}^{\mathrm{re}}
+
j\widehat{H}^{\mathrm{im}}.
\label{eq:cfft_complex}
\end{equation}

To enable learnable spectral interaction, CFFT applies a complex-valued linear transformation to the Fourier coefficients:
\begin{equation}
\begin{aligned}
\widetilde{H}^{\mathrm{re}}
&=
\phi
\left(
\widehat{H}^{\mathrm{re}}\mathbf{W}^{\mathrm{re}}
-
\widehat{H}^{\mathrm{im}}\mathbf{W}^{\mathrm{im}}
+
\mathbf{b}^{\mathrm{re}}
\right),\\
\widetilde{H}^{\mathrm{im}}
&=
\phi
\left(
\widehat{H}^{\mathrm{im}}\mathbf{W}^{\mathrm{re}}
+
\widehat{H}^{\mathrm{re}}\mathbf{W}^{\mathrm{im}}
+
\mathbf{b}^{\mathrm{im}}
\right),
\end{aligned}
\label{eq:cfft_interaction}
\end{equation}
where $\mathbf{W}^{\mathrm{re}}$, $\mathbf{W}^{\mathrm{im}}$, $\mathbf{b}^{\mathrm{re}}$, and $\mathbf{b}^{\mathrm{im}}$ are learnable parameters, and $\phi(\cdot)$ denotes a nonlinear activation. This complex transformation jointly updates amplitude- and phase-related channel-frequency responses. The refined spectrum is written as
\begin{equation}
\widetilde{H}
=
\widetilde{H}^{\mathrm{re}}
+
j\widetilde{H}^{\mathrm{im}}.
\label{eq:cfft_refined_complex}
\end{equation}

The feature representation is reconstructed by the inverse transform:
\begin{equation}
H'_{b,c,t}
=
\frac{1}{C}
\sum_{k=0}^{C-1}
\widetilde{H}_{b,k,t}
\exp\left(
j\frac{2\pi k c}{C}
\right),
\quad
c=0,\ldots,C-1,
\label{eq:cfft_idft}
\end{equation}

\section{Additional Module Details}\label{app:modules}

\subsection{Temporal Difference Computation in TDMM}
Given an RF feature sequence $I_f \in \mathbb{R}^{B \times C_i \times T_i}$, TDMM constructs a five-frame temporal neighborhood $\{X_{t-2},X_{t-1},X_t,X_{t+1},X_{t+2}\}$ and computes adjacent differences:
\begin{flalign}
& D_{-2} = X_{t-1}-X_{t-2}, \quad
D_{-1} = X_t-X_{t-1}, \notag\\
& D_{1}  = X_{t+1}-X_t, \quad
D_{2}  = X_{t+2}-X_{t+1}.
\label{eq:tdmm_difference}
\end{flalign}
The difference maps are concatenated and aggregated by a temporal convolution:
\begin{equation}
X_0 =
\operatorname{BN}
\left(
\operatorname{Conv}_{7 \times 1}
\left(
\operatorname{Concat}(D_{-2},D_{-1},D_{1},D_{2})
\right)
\right).
\label{eq:tdmm_conv}
\end{equation}

\subsection{Spatial Attention in SCFM}
Given $X_{\mathrm{fu}}$, a lightweight $5 \times 5$ convolutional stem generates a spatial attention response $A=\sigma(\operatorname{Stem}(X_{\mathrm{fu}}))$, which is normalized and applied to the feature map:
\begin{equation}
M =
\frac{(H'W')A}{2\|A\|_{1}+\epsilon},
\qquad
X_{\mathrm{attn}}=M\odot X_{\mathrm{fu}}.
\label{eq:scfm_attention}
\end{equation}

\subsection{Intermediate Refinement in RFAM}
RFAM first extracts local temporal patterns using a $7 \times 1$ convolution:
RFAM first extracts local temporal patterns using a $7 \times 1$ convolution:
\begin{equation}
X_r =
\operatorname{BN}
\left(
\operatorname{Conv}_{7 \times 1}(X)
\right).
\end{equation}
\noindent A channel attention vector is computed from average- and max-pooled temporal descriptors:
\begin{equation}
\begin{gathered}
s =
\sigma
\left(
\operatorname{MLP}(\operatorname{AvgPool}(X_r))
+
\operatorname{MLP}(\operatorname{MaxPool}(X_r))
\right), \\[0.7em]
\widetilde{X}_r = s \odot X_r.
\end{gathered}
\end{equation}

\subsection{Channel-Spectral Refinement in Overview}
Finally, the Channel-wise Fast Fourier Transform (CFFT) module performs spectral refinement along the feature-channel dimension. Instead of estimating physiological frequency directly from raw temporal signals, CFFT enhances informative inter-channel dependencies and suppresses redundant responses in the learned feature space:
\begin{equation}
H_{c}^{n_{5}}=\mathrm{CFFT}(H_{c}^{n_{4}}),\qquad
H_{f}^{n_{5}}=\mathrm{CFFT}(H_{f}^{n_{4}}).
\label{eq:cfft_feature}
\end{equation}

\section{Loss Function}\label{app:loss}
\subsection{Loss Function}
\vspace{-0.5em}
We train with the Negative Pearson Loss~\cite{yu2019remote}:
\begin{equation}
\mathcal{L}_{NP}(y,\hat{y}) = 1 - \rho(y,\hat{y}),
\end{equation}

\noindent where $\rho(\cdot,\cdot)$ denotes the Pearson correlation coefficient between the predicted and ground-truth physiological waveforms.

\end{document}